\pdfoutput=1

\documentclass[11pt]{article}

\usepackage[final]{acl}

\usepackage{times}
\usepackage{latexsym}

\usepackage[T1]{fontenc}

\usepackage[utf8]{inputenc}

\usepackage{microtype}

\usepackage{inconsolata}

\usepackage{graphicx}
\usepackage{xcolor}         
\usepackage{comment}
\newcommand\ours{\textit{Instruction Pre-Training}}
\newcommand\baseline{\textit{Vanilla Pre-Training}}
\usepackage{wrapfig}
\usepackage{subcaption}
\usepackage{caption}
\usepackage{array} 
\usepackage{booktabs} 
\usepackage{multirow} 
\usepackage{enumitem} 
\setlist[itemize]{noitemsep, topsep=0pt}
\usepackage[normalem]{ulem}
\useunder{\uline}{\ul}{}
\usepackage{amsmath}
\usepackage{marvosym}
\usepackage{url}

\title{Instruction Pre-Training:\\Language Models are Supervised Multitask Learners}

\author{
Daixuan Cheng$^{\dag}$\thanks{~Contribution during an internship at the CoAI Group, Tsinghua University. \textsuperscript{\Letter} Corresponding Author.}~Yuxian Gu$^{\ddag}$~Shaohan Huang$^{\dag}$\textsuperscript{\Letter}~~Junyu Bi$^{\dag}$~Minlie Huang$^{\ddag}$\textsuperscript{\Letter}~Furu Wei$^{\dag}$\\
$^\dag$ \small{Microsoft Research} \\ $^\ddag$ \small{The CoAI Group, DCST, Institute for Artificial Intelligence, State Key Lab of Intelligent Technology and Systems,}\\ \small{Beijing National Research Center for Information Science and Technology, 
Tsinghua University} \\
\small \texttt{daixuancheng6@gmail.com} \ \ \ \texttt{guyx21@mails.tsinghua.edu.cn} \\ \small \texttt{\{shaohanh,fuwei\}@microsoft.com}\ \ \ \texttt{aihuang@tsinghua.edu.cn}\\
\href{https://huggingface.co/instruction-pretrain}{https://huggingface.co/instruction-pretrain} 
}

\begin{document}
\maketitle
\begin{abstract}
Unsupervised multitask pre-training has been the critical method behind the recent success of language models (LMs). However, supervised multitask learning still holds significant promise, as scaling it in the post-training stage trends towards better generalization.
In this paper, we explore \textit{supervised multitask pre-training} by proposing~\ours, a framework that scalably augments massive raw corpora with instruction-response pairs to pre-train LMs. The instruction-response pairs are generated by an efficient instruction synthesizer built on open-source models. In our experiments, we synthesize 200M instruction-response pairs covering 40+ task categories to verify the effectiveness of~\ours. In pre-training from scratch, \ours~not only consistently enhances pre-trained base models but also benefits more from further instruction tuning. In continual pre-training,~\ours~enables Llama3-8B to be comparable to or even outperform Llama3-70B. 
Our model, code, and data are available at~\href{https://github.com/microsoft/LMOps}{https://github.com/microsoft/LMOps}.
\end{abstract}

\section{Introduction}
On the path towards general artificial intelligence, multitask learning~\citep{multitask} emerges as a promising approach. However, scaling supervised multitask learning to the necessary degree is very challenging. This motivates GPT-2~\citep{gpt2} to explore unsupervised multitask learning: pre-training on raw corpora through causal language modeling, which facilitates scaling up training data. Over time, unsupervised multitask learning has evolved into the standard approach for pre-training language models (LMs)~\citep{gpt3,palm}, which is referred to as \baseline~in this paper.

\begin{figure}[!tb]
    \centering
\includegraphics[width=\linewidth]{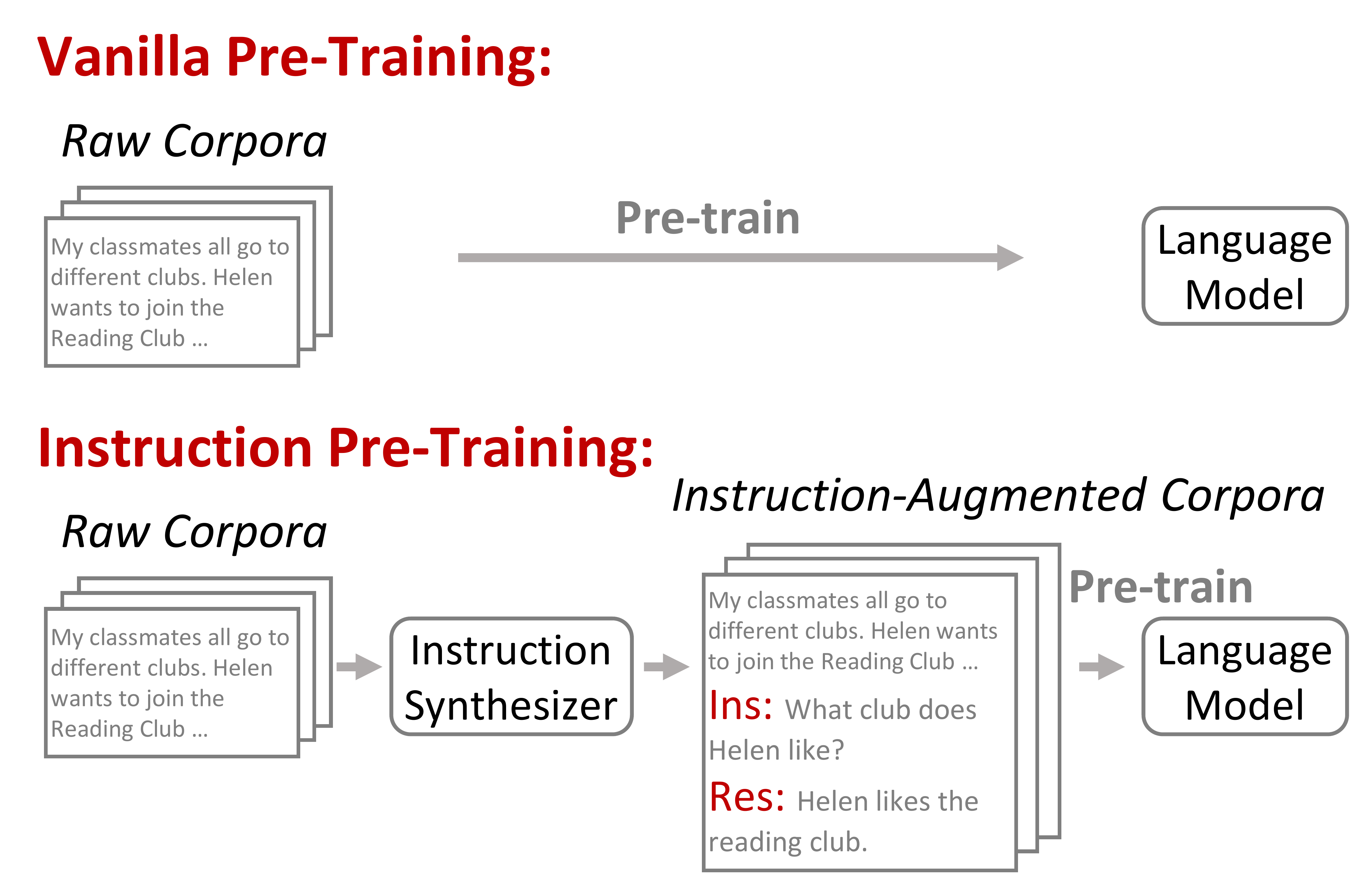}
    \caption{\textbf{Comparison between Instruction Pre-Training and Vanilla Pre-Training.} Instead of directly pre-training on raw corpora, \ours~augments the corpora with instruction-response pairs generated by an instruction synthesizer, then pre-trains LMs on the augmented corpora. ``Ins'' and ``Res'' represent instruction and response, respectively.}
    \label{fig:intro}
    \vspace{-10pt}
\end{figure}

Despite the success of unsupervised approaches, supervised multitask learning still holds significant promise. Instruction tuning~\citep{flan}, which fine-tunes pre-trained models using diverse tasks framed through natural language instructions, significantly enhances task generalization~\citep{t0,flanv2}, re-emphasizing the value of supervised multitask learning.

In this paper, we introduce \ours~to explore supervised multitask learning for pre-training. As shown in Figure~\ref{fig:intro}, instead of directly pre-training on raw corpora, \ours~augments each raw text with a set of instruction-response pairs\footnote{We use ``task'' and ``instruction-response pair'' interchangeably, with the instruction as task input and the response as task output.} generated by an instruction synthesizer, and then pre-trains LMs using the augmented corpora. These pairs are synthesized based on the content of massive raw corpora, ensuring high knowledge coverage and correctness. Therefore, we can scale up task synthesis with great diversity and quality~\citep{selfalignment}.

To develop the instruction synthesizer, we convert a wide range of existing datasets into our required format: each example consists of a set of instruction-response pairs and a piece of raw text that these pairs condition on.
Using this data collection, we fine-tune a language model to generate instruction-response pairs based the corresponding raw text. The high diversity of the tuning data enables the synthesizer to generalize to unseen data, facilitating the synthesis of instruction-response pairs for raw pre-training corpora. Unlike existing works~\citep{phi1.5,genie} using large or closed-source models~\citep{gpt4, genie} to generate synthetic data, we build our instruction synthesizer based on open-source models (typically with 7B parameters), which is much more cost-effective. This efficiency allows us to further scale up task synthesis: augmenting the raw corpora with 200M instruction-response pairs across more than 40 task categories.

We conduct experiments in both general pre-training from scratch and domain-adaptive continual pre-training. In pre-training from scratch, our 500M model pre-trained on 100B tokens reaches performance of the 1B model pre-trained on 300B tokens. Moreover, models that have undergone~\ours~gain significantly more from further instruction tuning. In continual pre-training, \ours~consistently improves performance of Llama3-8B\footnote{\href{https://llama.meta.com/llama3/}{https://llama.meta.com/llama3/}} on two domains: finance and biomedicine, enabling it to be comparable to or even surpass Llama3-70B.

In summary, our contributions include:
\begin{itemize}[leftmargin=*]
\itemsep0em 
\item We propose \ours~to explore supervised multitask pre-training, and verify its effectiveness through extensive experiments.
\item We develop an instruction synthesizer capable of scalably generating diverse instruction-response pairs based on various raw corpora.
\item We comprehensively analyze the instruction synthesizer and the synthetic data to reveal the key factors towards the success of our method.
\end{itemize}

\begin{figure*}[!t]
    \centering
    \includegraphics[width=\linewidth]{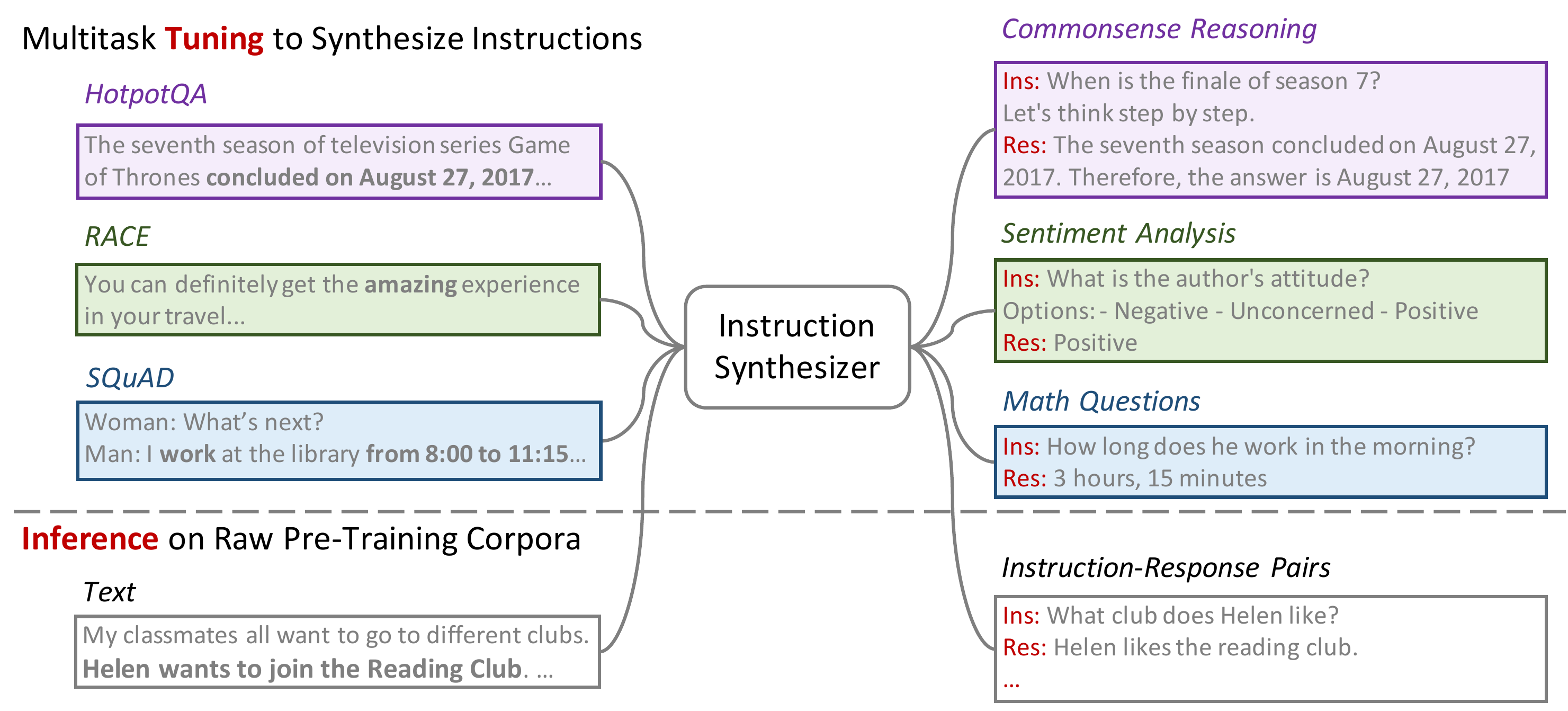}
    \caption{\textbf{Tuning and inference framework of instruction synthesizer.} During tuning, the instruction synthesizer learns to generate instruction-response pairs for a given raw text. The tuning data are curated to be highly diverse, enabling the synthesizer to generalize to unseen data. During inference, we use this tuned instruction synthesizer to generate instruction-response pairs for raw texts from pre-training corpora.}
    \label{fig:instruction_synthesizer}
\end{figure*}

\section{Instruction Pre-Training}
Instead of directly pre-training on raw corpora, \ours~augments each text from the raw corpora with a set of instruction-response pairs generated by an instruction synthesizer, where the instruction serves as the task input and the response serves as the task output, then pre-trains LMs on the augmented corpora.

\begin{figure}[!tb]
    \centering\includegraphics[width=\columnwidth]{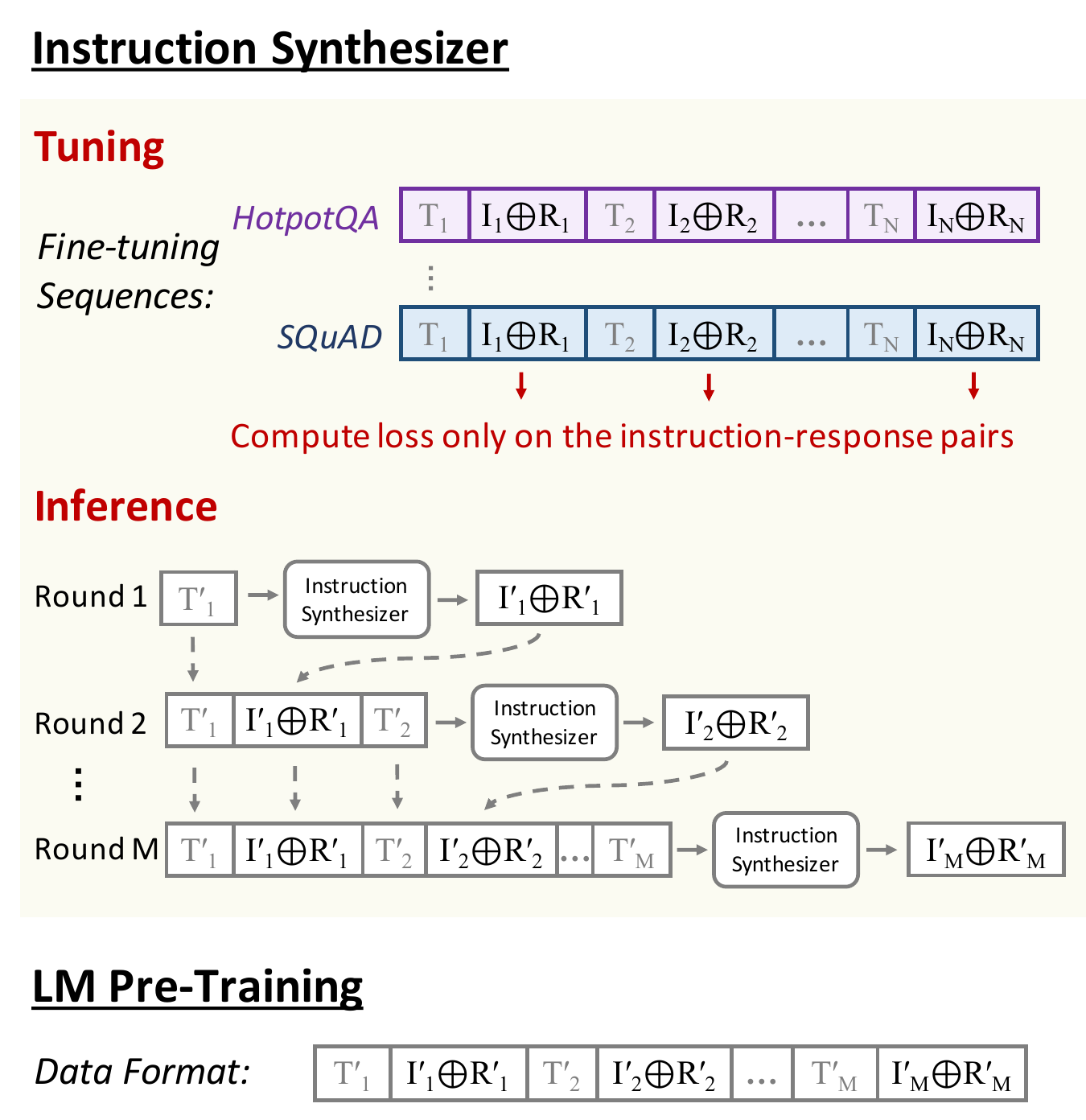}
    \caption{\textbf{For instruction synthesizer}, a one-shot example consists of a raw text (T$_\text{N}$) and a set of instruction-response pairs (I$_\text{N}$$\bigoplus$R$_\text{N}$); data denoted without $'$ are for tuning the instruction synthesizer, and data with $'$ are for synthesizer inference and LM pre-training. During instruction synthesizer tuning, each sequence fed into the synthesizer concatenates multiple one-shot examples sampled from the same dataset. During inference, multi-round inference is conducted to synthesize instruction-response pairs with patterns similar to those of previous rounds. \textbf{For LM pre-training}, a few-shot example concatenates raw texts and synthesized pairs from multiple rounds.}
\vspace{-20pt}
\label{fig:train_inference_details}
\end{figure}
\subsection{Instruction Synthesizer}
To facilitate the scaling of supervised task learning, we develop an instruction synthesizer to generate instruction-response pairs based on raw corpora. 
Studies suggest that raw corpora contain numerous intrinsic tasks~\citep{udit,skillit}, which enables efficient scaling of task synthesis~\citep{adaptllm,selfalignment,mammoth2} along with the upscale of raw corpora. 

Our instruction synthesizer is developed through multitask fine-tuning on a language model. As illustrated in Figure~\ref{fig:instruction_synthesizer}, during tuning, the instruction synthesizer is given a piece of raw text and tuned to create a set of instruction-response pairs. 
The tuning data are curated to be highly diverse, enabling the instruction synthesizer to generalize to unseen data~\citep{flan}. Therefore, during inference, we can directly employ the instruction synthesizer to create instruction-response pairs based on the raw pre-training corpora. Furthermore, we incorporate specific designs to synthesize both one-shot and few-shot examples for subsequent pre-training.

\paragraph{Data Collection} We sample from and reformat a diverse range of context-based task completion datasets, which require models to perform tasks based on a given context, to meet our fine-tuning requirements.
Each data sample's context serves as the raw text, and the downstream tasks serve as the instruction-response pairs. The contexts span various domains such as encyclopedias, social media, and academic tests~\citep{QAsurvey}, and the tasks encompass a wide range such as commonsense reasoning and sentiment analysis. Further details are in Appendix~\ref{sec:Fine-Tuning Data Collection}.

\paragraph{Tuning} We tune the instruction synthesizer using few-shot examples. 
As depicted in Figure~\ref{fig:train_inference_details}, a one-shot example consists of a piece of raw text followed by its instruction-response pairs. 
Each sequence fed into the synthesizer concatenates multiple such examples, all sampled from the same dataset. 
This ensures that the concatenation of multiple examples within one sequence constitutes a few-shot example, maintaining consistency in patterns (i.e., task format or category) among different sets of instruction-response pairs. 
Fine-tuning on these examples enables the instruction synthesizer to generate instruction-response pairs with similar patterns to those in the given examples~\citep{metaicl}. 
Additionally, we calculate the tuning loss only on the instruction-response pairs to guide the model to focus on these pairs.

\paragraph{Inference} We conduct multi-round inference to create few-shot examples. 
As depicted in Figure~\ref{fig:train_inference_details}, in each round, we prepend the texts and instruction-response pairs from previous rounds to the current text. This allows the instruction synthesizer to generate new instruction-response pairs based on the previous ones.

\subsection{LM Pre-Training}
After collecting the synthesized instruction-response pairs, we employ templates from~\citet{flancollection} to diversify instruction formats, and templates from~\citet{adaptllm} to concatenate each raw text with its instruction-response pairs. 
As shown in Figure~\ref{fig:train_inference_details}, by concatenating the texts and instruction-response pairs from $M$ rounds, we create an $M$-shot example for subsequent pre-training.

Except for the pre-training data, \ours~keeps all other pre-training settings the same as \baseline: training with the next-token prediction objective~\citep{gpt1} and computing loss on all tokens. We conduct both general pre-training from scratch and domain-adaptive continual pre-training to verify the effectiveness in different pre-training scenarios.

\paragraph{General Pre-Training From Scratch}
Considering the large amount of data required for general pre-training from scratch, we only convert part of the raw corpora into instruction-augmented corpora, leaving the rest unchanged. Besides, we mix the corpora with the data for fine-tuning the instruction synthesizer to enhance task diversity. 

\paragraph{Domain-Adaptive Continual Pre-Training}
For domain-adaptive continual pre-training, the data requirement is much smaller. Therefore, we convert all raw corpora into instruction-augmented corpora. Following~\citet{adaptllm}, we mix the corpora with the general instructions to benefit from improved prompting ability. Since the general instructions collection contains the fine-tuning data for the instruction synthesizer, we do not include these fine-tuning data.
\begin{table*}[!htb]
\centering
\small
\resizebox{\linewidth}{!}{%
\begin{tabular}{lccccccccc}
\toprule
                   & ARC-e              & ARC-c           & BoolQ                   & SIQA            & WinoGrande      & PIQA            & OBQA            & HellaSwag       & MMLU            \\ \midrule
\multicolumn{10}{l}{\hspace{-0.22cm} {\ul \textit{500M}}}                                                                                                                      \\\vspace{0.09cm}
Vanilla PT         & 50.3            & 26.4            & 57.5                     & 44.6            & 53.8            & \textbf{71.1}            & 29.8            & 47.2            & 25.4            \\
Mix PT & 52.8            & 26.7            & 46.8                       & 46.6            & 52.7            & 70.1   & 30.0            & 47.0            & \textbf{26.7}   \\
Instruct PT        & \textbf{54.8}   & \textbf{27.4}   & \textbf{62.0}     & \textbf{47.2}   & \textbf{54.8}   & 69.9            & \textbf{30.8}   & \textbf{47.3}   & 25.3            \\ \midrule
\multicolumn{10}{l}{\hspace{-0.22cm} {\ul \textit{1.3B}}}                                                                                                                      \\\vspace{0.09cm}
Vanilla PT         & 58.5            & 28.8            & 60.3                   & 47.9            & 54.9            & 73.0           & \textbf{33.6}   & \textbf{54.9}   & 25.7            \\
Instruct PT        & \textbf{60.5}   & \textbf{30.9}   & \textbf{62.2}     & \textbf{49.2}   & \textbf{55.9}   & \textbf{73.6}   & 33.4            & 54.3            & \textbf{27.3}  \\ \bottomrule
\end{tabular}%
}
\caption{\textbf{General performance of the pre-trained base models} via \baseline~(Vanilla PT), mixing raw corpora with fine-tuning data for the instruction synthesizer (Mix PT), and \ours~(Instruct PT) in general pre-training from scratch. All the pre-training methods use the same number of tokens for model training.}
\label{tab:general_performance}
\end{table*}

\begin{table}[!t]
\centering
\small
\resizebox{0.95\linewidth}{!}{%
\begin{tabular}{lllc}
\toprule
            & \# Param. & \# Token & Average \\\midrule
GPT-2 & 774M      & -         & 45.7 \\
Pythia      & 1B        & 300B      & 47.1 \\
BLOOM       & 1.1B      & 341B         & 45.1 \\\midrule[0.05pt]
Instruct PT & 500M      & 100B      & 46.6 \\
\midrule
OPT         & 1.3B      & 300B      & 49.3 \\
GPT-2    & 1.5B      & -         & 48.6  \\
BLOOM      & 3B      & 341B      & 50.1 \\ \midrule[0.05pt]
Instruct PT & 1.3B      & 100B      & 49.7  \\
\bottomrule
\end{tabular}%
}
\caption{\textbf{Comparison between our pre-trained base models and others} on general benchmarks. Detailed results are in Table~\ref{tab:general_comprison}.}
\label{tab:avg_general_comprison}
\vspace{-5pt}
\end{table}

\section{Experiment Settings}
\subsection{Instruction Synthesizer}
Our synthesizer is fine-tuned from Mistral-7B-v0.1~\citep{mistral}, an open-source model with 7B parameters. This model is much more cost-effective than large-scale~\citep{falcon,Mixtral,Qwen} or closed-source~\citep{gpt4} models typically used for generating synthetic data~\citep{phi1.5,genie,mammoth2}. During inference, about 5 instruction-response pairs are created per raw text, where each pair contains about 52 tokens. Further tuning and inference details are in Appendix~\ref{sec:Tuning and Inference Settings for Instruction Synthesizer}.

\subsection{General Pre-training From Scratch}
\paragraph{Pre-Training Corpora} We randomly sample a subset of RefinedWeb~\citep{refinedweb} dataset for raw pre-training corpora, consisting of 200M pieces of text containing about 100B tokens. 

To create instruction-augmented corpora, we conduct two rounds of instruction synthesis, converting 1/5 of the raw corpora (40M raw texts) into instruction-augmented texts. The first round converts 20M raw texts, and the second round uses the raw texts and instruction-response pairs from the first round to convert another 20M raw texts. The resulted corpora contain 200M synthesized pairs amounting to about 10B tokens.
An example of a 2-shot instruction-augmented text is shown in Table~\ref{tab:pt_case_general} in Appendix.

We then mix the fine-tuning data for instruction synthesizer. Since the fine-tuning data amount (0.2B tokens) is too small compared to that of the raw corpora, we increase its sample ratio so that it repeats 4 times throughout pre-training.

\paragraph{Training and Evaluation} We adopt the architecture and tokenizer of Mistral~\citep{mistral} to implement models of two different parameters: 500M and 1.3B. 

Our pre-training settings largely follow \citet{gpt3}. To enhance training efficiency, we implement the memory-efficient attention of \textit{xformers}~\citep{xFormers}. Detailed hyperparameters are listed in Table~\ref{tab:Hyper-parameters of General Pre-training From Scratch} in Appendix. The lm-evaluation-harness framework~\citep{harness} is used for model evaluation, detailed evaluation settings are in Appendix~\ref{sec:Language Model Evaluation Setting}.

We also conduct instruction tuning on the pre-trained model with 500M parameters using the data from~\citet{flancollection}. 
The instruction-tuned models are evaluated on MMLU~\citep{mmlu} benchmark.

\subsection{Domain-Adaptive Continual Pre-Training}
\paragraph{Pre-Training Corpora} We use raw corpora from two domains: PubMed Abstracts~\citep{Pile} for biomedicine and financial news~\citep{fingpt} for finance.

We conduct 3-round inference to covert all the domain-specific corpora. Each round processes 1/3 of the raw corpora, inheriting the raw texts and instruction-response pairs from previous rounds. Examples of the instruction-augmented texts are in Table~\ref{tab:pt_case_med} and~\ref{tab:pt_case_finance} in Appendix.

We then mix the instruction-augmented corpora with general instructions~\citep{lima,wizardlm,OpenOrca}, using the same mixing ratio as~\citet{adaptllm}.

\paragraph{Training and Evaluation} We continue to pre-train Llama3-8B on each domain respectively, detailed settings are in Table~\ref{tab:Hyper-parameters of General Pre-training From Scratch} in Appendix. We follow the prompting settings in~\citet{adaptllm} to evaluate models on the domain-specific tasks. Detailed evaluation settings are in Appendix~\ref{sec:Language Model Evaluation Setting}.

\section{Results}
\subsection{General Pre-Training From Scratch}

\paragraph{Pre-Trained Base Models} Table~\ref{tab:general_performance} presents the general performance of the models after pre-training. To ensure a fair comparison with \baseline, which uses only raw corpora, we include a baseline (Mix PT) that mixes the raw corpora with the fine-tuning data for our instruction synthesizer. Compared to \baseline~(Vanilla PT), incorporating the fine-tuning data in Mix PT improves model performance on several benchmarks. By further transforming the raw corpora into instruction-augmented corpora, \ours~(Instruct PT) achieves even better performance. Note that none of the evaluated datasets are included in our fine-tuning data for the instruction synthesizer. Nevertheless, the model pre-trained on the data generated by the instruction synthesizer shows improved performance on these unseen datasets, demonstrating the effectiveness of our method in enhancing model generalization.

In Table~\ref{tab:avg_general_comprison}, we compare our pre-trained models with other open-source models. Using 100B tokens, our 500M model reaches the performance of Pythia-1B~\citep{pythia} trained with 300B tokens and our 1.3 B model reaches the performance of BLOOM-3B~\citep{Bloom} trained with 341B tokens. This shows consistent data efficiency of~\ours~across different model scales.
\begin{figure}[!tb]
    \centering
    \includegraphics[width=\linewidth]{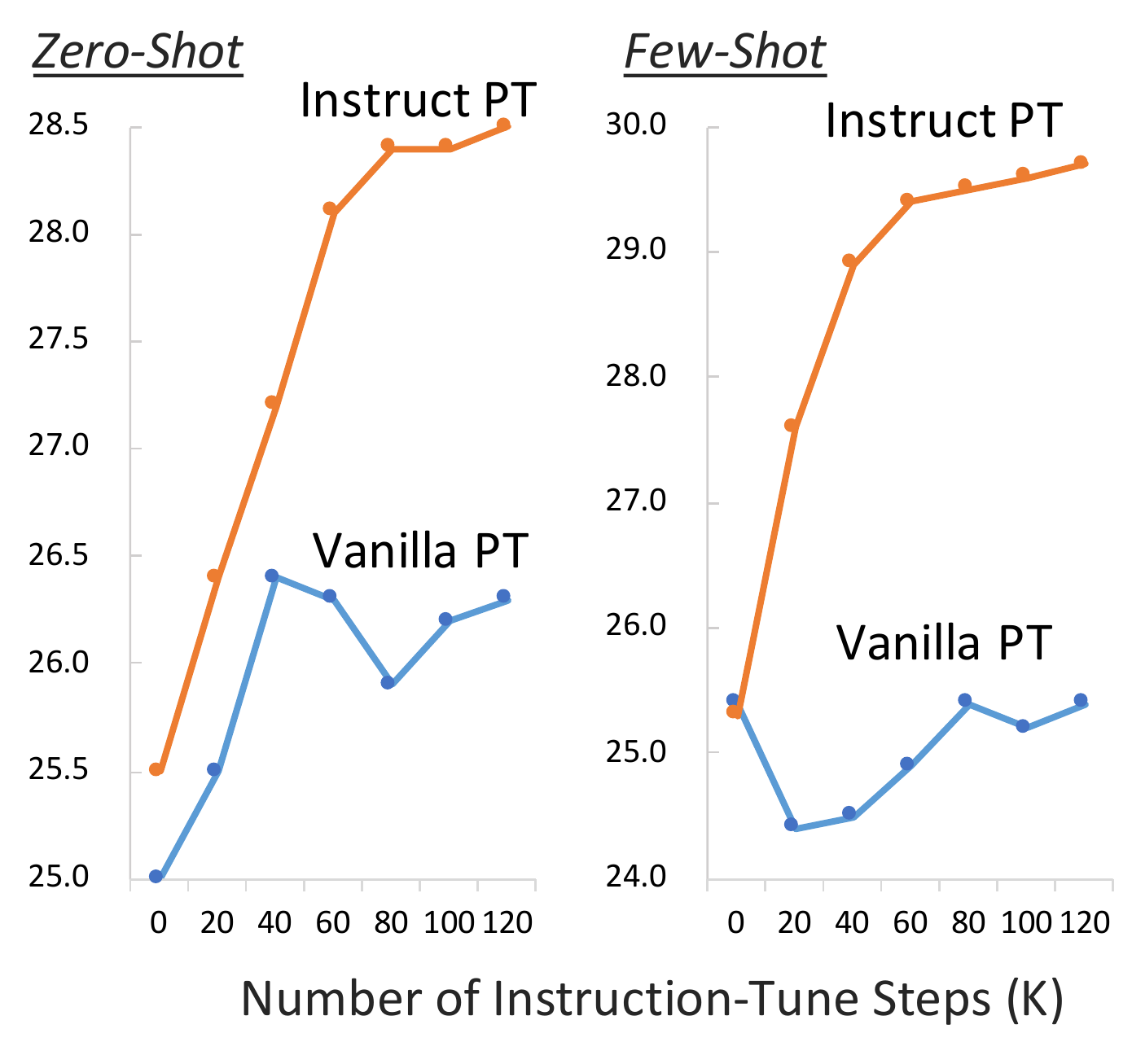}
    \vspace{-15pt}
    \caption{\textbf{MMLU performance during instruction tuning} of models pre-trained via \baseline~(Vanilla PT) and \ours~(Instruct PT).}
    \vspace{-10pt}
\label{fig:mmlu}
\end{figure}

\begin{table*}[!tb]
\centering
\begin{tabular}{lccccc|c}
\toprule
\textit{BioMed.} & PubMedQA              & ChemProt           & RCT              & MQP                & UMSLE           & \textsc{Average}            \\ \midrule
\textcolor{gray}{Llama3-70B} & \textcolor{gray}{54.3}                          & \textcolor{gray}{51.8}                        & \textcolor{gray}{82.2}                       & \textcolor{gray}{84.8}                        & \textcolor{gray}{46.7}                        & \textcolor{gray}{63.9}                        \\
Llama3-8B         & 59.8                         & 27.6                       & \textbf{73.6}                         &  66.2                      & \textbf{40.6}                        & 53.6                        \\
Vanilla PT-8B    & 65.1                        &  42.4            & 72.4              & 76.4                          & 35.5               & 58.4                        \\

Instruct PT-8B     & \textbf{68.7}               & \textbf{47.2}               &   73.4                      & \textbf{79.3}            & 38.0                       & \textbf{61.3} 
\\\bottomrule
\end{tabular}
\vspace{3mm}

\begin{tabular}{lccccc|c}
\toprule
\textit{Finance}  & ConvFinQA             & Headline                 & FiQA SA   & FPB            & NER           & \textsc{Average}             \\ \midrule
\textcolor{gray}{Llama3-70B} & \textcolor{gray}{59.1}        & \textcolor{gray}{86.3}                        & \textcolor{gray}{81.0}         & \textcolor{gray}{68.5} & \textcolor{gray}{64.4} & \textcolor{gray}{71.9} \\ 
Llama3-8B   & 49.9                        & 81.1                         & \textbf{83.3}                 & 63.5                          & \textbf{72.8}                         & 70.1                         \\
Vanilla PT-8B   & 62.9                        & 84.7                         &  82.2                 & 65.4                          & 64.9                         & 72.0                         \\
Instruct PT-8B      & \textbf{74.6}                & \textbf{87.1}               &  82.4               &\textbf{65.7}                 & 63.6                & \textbf{74.7}                \\ 
 \bottomrule
\end{tabular}

\caption{\textbf{Domain-specific task performance} of Llama3-8B without continued pre-training, after continued pre-training via \baseline~(Vanilla PT), and after continued pre-training via \ours~(Instruct PT). Both Vanilla PT and Instruct PT mix domain-specific corpora with general instructions to boost prompting ability, and use the same number of tokens for model training. The performance of Llama3-70B is displayed for reference.}
\vspace{-5pt}
\label{tab:domain_results}
\end{table*}

\begin{table}[!tb]
\centering
\resizebox{\linewidth}{!}{%
\begin{tabular}{lcccc}
\toprule
 & w/o Corpora & Rule-based & 1-shot & Ours \\ \midrule
Med.  & 58.6 & 58.8  & 58.5               & \textbf{61.3}     \\ 
Fin.   & 73.3 & 73.1  & 73.1                  & \textbf{74.7}     \\\bottomrule

\end{tabular}
}
\caption{\textbf{Ablations on training data.} \textit{w/o Corpora} removes domain-specific pre-training corpora. \textit{Rule-based} replaces instruction-augmented corpora with those created by the rule-based methods in \citet{adaptllm}. \textit{1-shot} replaces instruction-augmented corpora with those created through single-turn synthesis. We report the average task scores within each domain.}
\vspace{-5pt}
\label{tab:domain_ablation}
\end{table}

\paragraph{Instruction-Tuned Models}
Figure~\ref{fig:mmlu} shows the zero/few-shot performance on MMLU during instruction tuning from the pre-trained models. The model pre-trained via~\ours~quickly outperforms the model pre-trained via~\baseline, and we observe a stable increasing trend of our model throughout the instruction tuning process. We infer that the closer alignment of training tasks during the instruction pre-training and instruction tuning stages facilitates a smoother transition between pre-training and fine-tuning. This alignment enables the model to learn more rapidly on downstream tasks. Therefore, \ours~offers a promising solution to significantly reduce the number of further fine-tuning steps~\citep{flancollection,pit}.

\subsection{Domain-Adaptive Continual Pre-Training}
\paragraph{Main Results} As shown in Table~\ref{tab:domain_results}, \ours~consistently outperforms \baseline~on almost all domain-specific tasks. Continual pre-training with \ours~significantly enhances the domain-specific performance of Llama3-8B, achieving parity with or even surpassing Llama3-70B. On the finance NER benchmark, where \ours~underperforms~\baseline, we observe considerable variance, where even Llama3-70B underperforms Llama3-8B, suggesting that this benchmark may not be reliable.

\paragraph{Ablations} Table~\ref{tab:domain_ablation} presents ablation results for our pre-training data, which consist of a mixture of domain-specific instruction-augmented corpora and general instructions.

\begin{itemize}[leftmargin=*]
\itemsep0em 
\item \textit{w/o Corpora}: Removing the domain-specific instruction-augmented corpora eliminates the source of domain-specific knowledge, leading to reduced domain-specific performance.
\item \textit{Rule-based}: Constructing instruction-augmented corpora using rule-based methods results in limited diversity, thereby constraining performance.
\item \textit{1-shot}: Limiting synthesis to 1-turn instead of multi-turn synthesis results in instruction-augmented corpora containing only 1-shot examples, leading to decreased prompting performance~\citep{flancollection}.
\end{itemize}

\section{Analysis}\label{sec:analysis}
We conduct a detailed analysis of the instruction synthesizer and the instruction-augmented corpora to understand their impact on LM pre-training.

\subsection{Instruction Synthesizer}
Our goal in multitask fine-tuning is to develop a general synthesizer capable of generating instruction-response pairs for any raw text. Therefore, we evaluate its performance on both seen datasets (listed in Appendix~\ref{sec:Fine-Tuning Data Collection}) and unseen datasets. The unseen datasets include SocialIQA~\citep{siqa}, TextbookQA~\citep{textbookqa}, WikiWhy~\citep{wikiwhy}, and FEVER~\citep{fever}, each representing a specific instruction format. Each example in these datasets comprises a context (raw text) and a set of context-based tasks (instruction-response pairs).
\begin{table}[!tb]
\centering
\small
\resizebox{\linewidth}{!}{%
\begin{tabular}{lcccccc}
\toprule
\multirow{3}{*}{} & \multicolumn{2}{c}{Accuracy}  & \multicolumn{4}{c}{Quality}         \\ \cmidrule(r){2-3} \cmidrule(r){4-7} 
                  & \multirow{2}{*}{Seen} & \multirow{2}{*}{Unseen} & \multicolumn{2}{c}{Seen}      & \multicolumn{2}{c}{Unseen}    \\ \cmidrule(r){4-5} \cmidrule(r){6-7} 
                  &                       &                         & Zero     & Few      & Zero     & Few      \\ \midrule
Base              & 30.6                  & 29.2                    & 16.5          & 21.8          & 12.1          & 19.6          \\
Ours              & \textbf{70.0}         & \textbf{55.2}           & \textbf{49.4} & \textbf{49.9} & \textbf{25.3} & \textbf{30.8} \\ \bottomrule
\end{tabular}%
}
\caption{\textbf{Response accuracy and instruction-response pair quality} of our instruction synthesizer (Ours) and Mistral-7B (Base). ``Zero'' indicates the zero-shot setting where no examples are presented before the testing raw text, and ``Few'' prepends 3-shot examples to the testing raw text.}
\label{tab:qulaity_acc}
\vspace{-10pt}
\end{table}

\paragraph{Response Accuracy} Given a raw text and a task instruction, the instruction synthesizer generates a response. We compute the F1 similarity between the generated response and the gold response to evaluate response accuracy. Our instruction synthesizer is fine-tuned from the base Mistral-7B model. For comparison, we also present the results of the base model. As shown in Table~\ref{tab:qulaity_acc}, our fine-tuned synthesizer significantly outperforms the base model on both seen and unseen datasets, demonstrating the effectiveness of our fine-tuning.

\paragraph{Instruction-Response Pair Quality} Given a raw text, the instruction synthesizer generates a set of instruction-response pairs. We compute the F1 similarity between the generated pairs and the gold pairs to evaluate their quality. The evaluation is conducted in both zero-shot and few-shot settings: 1) Zero-shot: the input to the instruction synthesizer contains only the raw text. 2) Few-shot: following~\citet{selfinstruct, genie}, a few examples from the same dataset as the gold instruction-response pairs, each consisting of a raw text and corresponding instruction-response pairs, are prepended to the testing raw text.

As shown in Table~\ref{tab:qulaity_acc}, compared to the base model, our fine-tuned synthesizer significantly outperforms the baseline across all four dimensions: zero-shot, few-shot, seen, and unseen datasets. In unseen datasets, the few-shot setting substantially outperforms the zero-shot setting, indicating that our synthesizer effectively leverages the pattern of the few-shot examples to create instruction-response pairs for the testing text. 

\paragraph{Helpfulness on LM Generalization}
We conduct experiments using an LM (base Mistral-7B in our analysis) to assess the impact of synthesized instruction-response pairs on helping LMs generalize to unseen tasks. Given a prompt concatenating a testing raw text, synthesized pairs, and a testing instruction, the LM generates a response. We then compare the LM's performance on the testing task with and without the synthesized pairs in the prompt to evaluate their effectiveness.

We evaluate instruction-response pairs generated using different methods: 1) Random: randomly sampled instruction-response pairs of a different context. 2) Base: pairs synthesized based on the testing raw text by the base Mistral-7B model prompted with a few examples. 3) Ours: pairs synthesized based on the testing raw text by our instruction synthesizer using the same few-shot examples as Base.

\begin{figure}[!tb]
    \centering
    \includegraphics[width=0.95\linewidth]{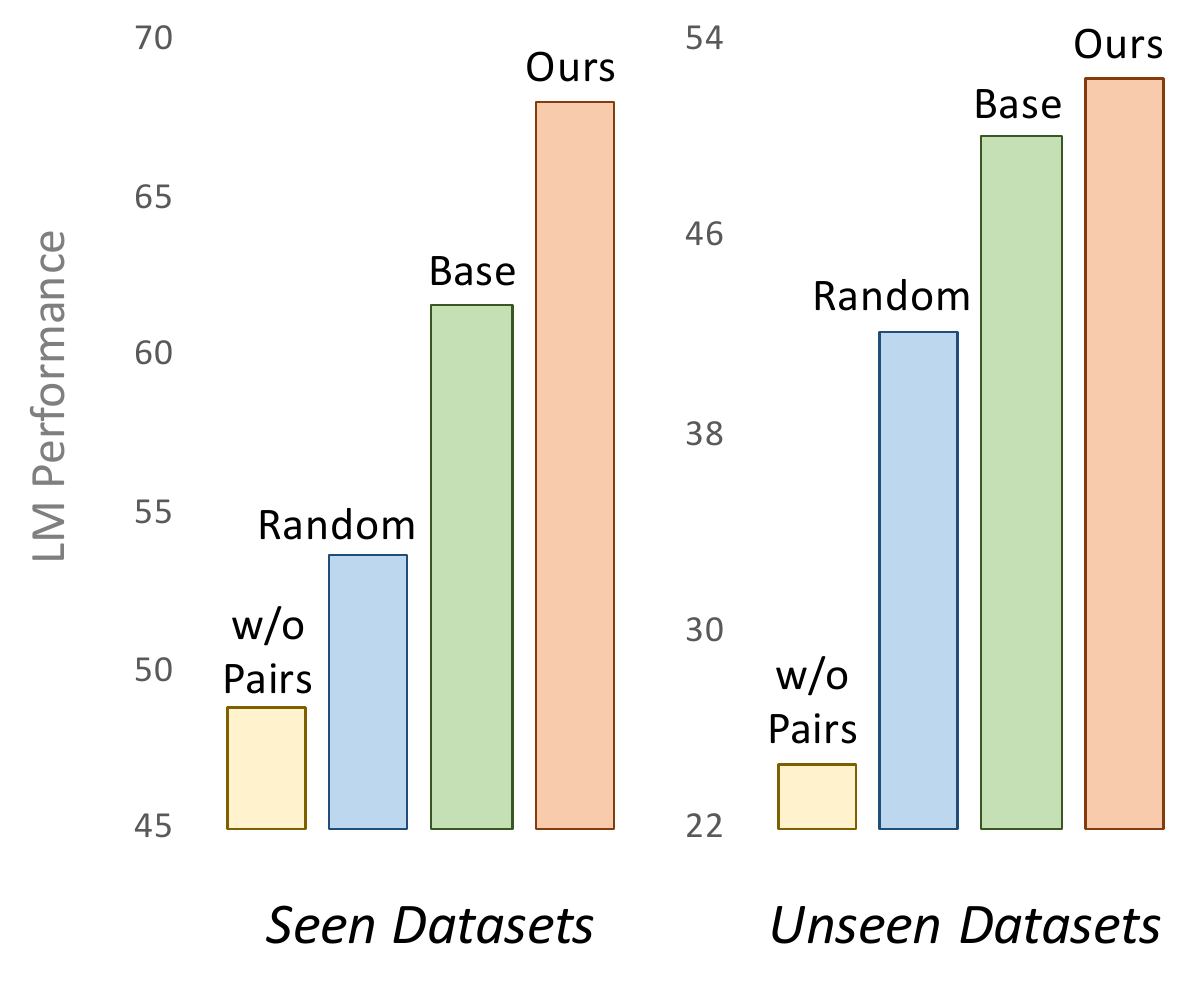}
    \vspace{-10pt}
    \caption{\textbf{Helpfulness on LM generalization} measured by LM performance with or without synthesized instruction-response pairs in the prompt.} 
\label{fig:helpfulness}
\end{figure}

As shown Figure~\ref{fig:helpfulness}, ``w/o Pairs'' denotes the setting where synthesized pairs are excluded from the prompt. On both seen and unseen datasets, ours consistently enhances the LM's performance on the testing task, surpassing all baselines. This demonstrates the effectiveness of our synthesized tasks in improving the LM's ability to perform a wide range of tasks.

\subsection{Instruction-Augmented Corpora}
We analyze the instruction-augmented pre-training corpora in terms of context relevance, response accuracy and task diversity. We sample 500 instruction-augmented texts from the augmented corpora and use GPT-4~\citep{gpt4} to evaluate the synthesized instruction-response pairs. Specifically, GPT-4 is prompted to assess whether the synthesized instruction is relevant to the context of the raw text (context relevance) and whether the response is accurate based on the instruction and context (response accuracy). Additionally, to evaluate task diversity, we prompt GPT-4 to categorize each instruction-response pair using a predefined list of task categories from~\citet{supernaturalinstructions}.
\begin{table}[!tb]
\centering
\small
\resizebox{\columnwidth}{!}{
\begin{tabular}{lccc}
\toprule
 & Accuracy  & Relevance & \# Category \\ \midrule
General   & 77.5 & 92.9  & 49  \\
BioMed.   & 86.2 & 99.4  & 26  \\
Finance   & 69.8 & 85.8  & 41  \\ \bottomrule
\end{tabular}
}
\caption{\textbf{Response accuracy, context relevance, and number of task categories} of the instruction-augmented corpora.}
\vspace{-5pt}

\label{tab:analysis}
\end{table}

\begin{figure}[!tb]
\centering
\includegraphics[width=0.9\linewidth]{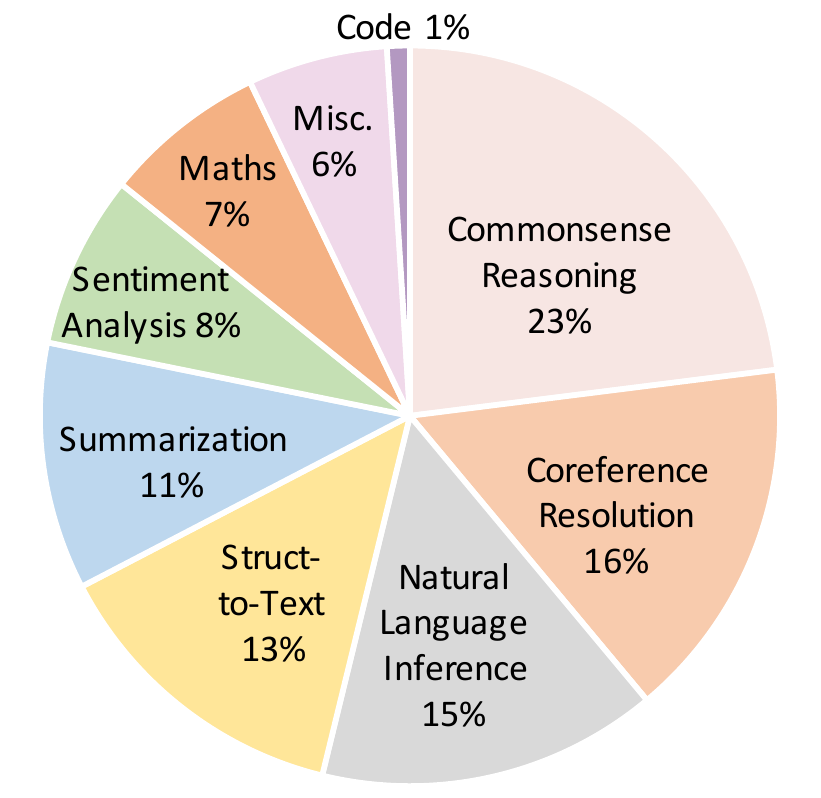}    
\caption{\textbf{Distribution of task scenarios of synthesized instruction-response pairs} in the instruction-augmented corpora.}
\label{fig:analysis}
\vspace{-5pt}
\end{figure}
As shown in Table~\ref{tab:analysis}, our instruction synthesizer generates instruction-response pairs spanning 49 different task categories, with over 85\% relevance to the context and 70\% response accuracy. We further group the task categories into 9 general task scenarios. Figure~\ref{fig:analysis} shows the percentages of each task scenario in the instruction augmented corpora for general pre-training. Our synthesized tasks cover all general task scenarios, demonstrating the effectiveness of our instruction synthesizer in generating a highly diverse tasks. We conduct further analysis in Appendix~\ref{sec: Human Evaluation on Instruction-Augmented Corpora} for human evaluation, Appendix~\ref{sec: Data Contamination Analysis} for data contamination, and Appendix~\ref{sec:Analysis on Domain Diversity} for domain distribution and diversity.

\section{Related Work}

\paragraph{Synthetic Instruction Generation}
There have been many works studying synthetic instruction generation, but they mainly focus on post-training~\citep{wizardlm,selfalignment}, while we focus on pre-training. This makes these works complementary to ours. Moreover, our experiments demonstrate that instruction pre-trained models gain more from instruction post-training, highlighting the complementary nature.

Regardless of the training stage, our method differs from related works in several ways. Firstly, we focus on learning from the raw corpora rather than distilling knowledge from strong models~\citep{wizardlm,orca,GLAN}. Secondly, ours can be task-agnostic, in contrast to the more task-specific approaches~\citep{selfinstruct,unnaturalinstructions,genie} relying on a few gold examples. Additionally, we outperforms rule-based methods~\citep{adaptllm,udit} by increasing instruction diversity. Moreover, the iterative techniques used in~\citet{selfalignment,llm2llm,mammoth2} could potentially complement our method, areas we plan to explore in future research.

\paragraph{Data Curation for LM Pre-Training}
Data curation for LM pre-training typically involves collection, cleaning, and organization. Most pre-training data are collected from the Internet to ensure diversity~\citep{C4,refinedweb,ccnet,Pile}.
Although diverse, web-scraped data often contain low-quality and duplicate content. Therefore, data cleaning techniques are applied to these corpora, including language identification~\citep{fasttext}, perplexity-based~\citep{ccnet}, classifier-based~\citep{gpt3}, and rule-based~\citep{C4,massiveweb} filtering. Data organization aims at performing more fine-grained programming of the data, including data selection~\cite{data_selection_survey,doremi} and constructing training instances related to downstream usage~\cite{ppt,picl,incontextpretrain,adaptllm1.5,rephrasetheweb}.
Our work explores an orthogonal direction: augmenting raw corpora with large-scale supervised signals.

\section{Conclusion}
This paper proposes \ours~to explore supervised multitask learning for pre-training. Instead of directly pre-training on raw corpora, \ours~augments the corpora with instruction-response pairs generated by an instruction synthesizer. Our instruction synthesizer, fine-tuned from a highly diverse data collection, is capable of generating diverse instruction-response pairs from various corpora. In pre-training from scratch, \ours~not only outperforms \baseline~on the pre-trained base models but also benefits more from further instruction tuning. In continual pre-training, \ours~substantially enhances the performance of Llama3-8B in two different domains. Looking ahead, we hope our work can inspire further exploration into this promising area of supervised multitask pre-training, effectively enhancing the general abilities of LMs.

\section*{Limitations}
While synthetic data offer numerous benefits, it is crucial to acknowledge the potential limitations. Our work, along with other works utilizing synthetic data~\citep{syntheticsurvey}, is inevitably limited by the possibility of introducing hallucinations. As shown in our analysis in Section~\ref{sec:analysis}, the accuracy of our instruction-augmented corpora is approximately 70\%, which may potentially mislead the pre-trained model. Future work could explore post-verification techniques such as those proposed by~\citet{selfalignment, llm2llm, mammoth2, genie} to filter out low-quality data or develop methods to enhance the reliability of the instruction synthesizer.

Furthermore, works like~\citet{llama2, mistral} have achieved impressive performance by pre-training on trillions of tokens, whereas our pre-training is currently limited to the scale of billions of tokens. Future research should investigate scaling laws for synthetic data and determine the optimal balance between quantity and quality of synthetic samples~\citep{syntheticsurvey}.

\section*{Ethics Statement}
Except for the pre-training corpora in the finance domain, all datasets and language models used in this work are publicly available.

\section*{Acknowledgements}
This work was supported by the National Science Foundation for Distinguished Young Scholars (with No. 62125604) and the NSFC key project (with No. 61936010).

\bibliography{custom,ref}

\clearpage
\appendix
\section{Data Collection for Fine-Tuning Instruction Synthesizer}\label{sec:Fine-Tuning Data Collection}
\begin{figure*}[!htb]
    \centering
    \includegraphics[width=\linewidth]{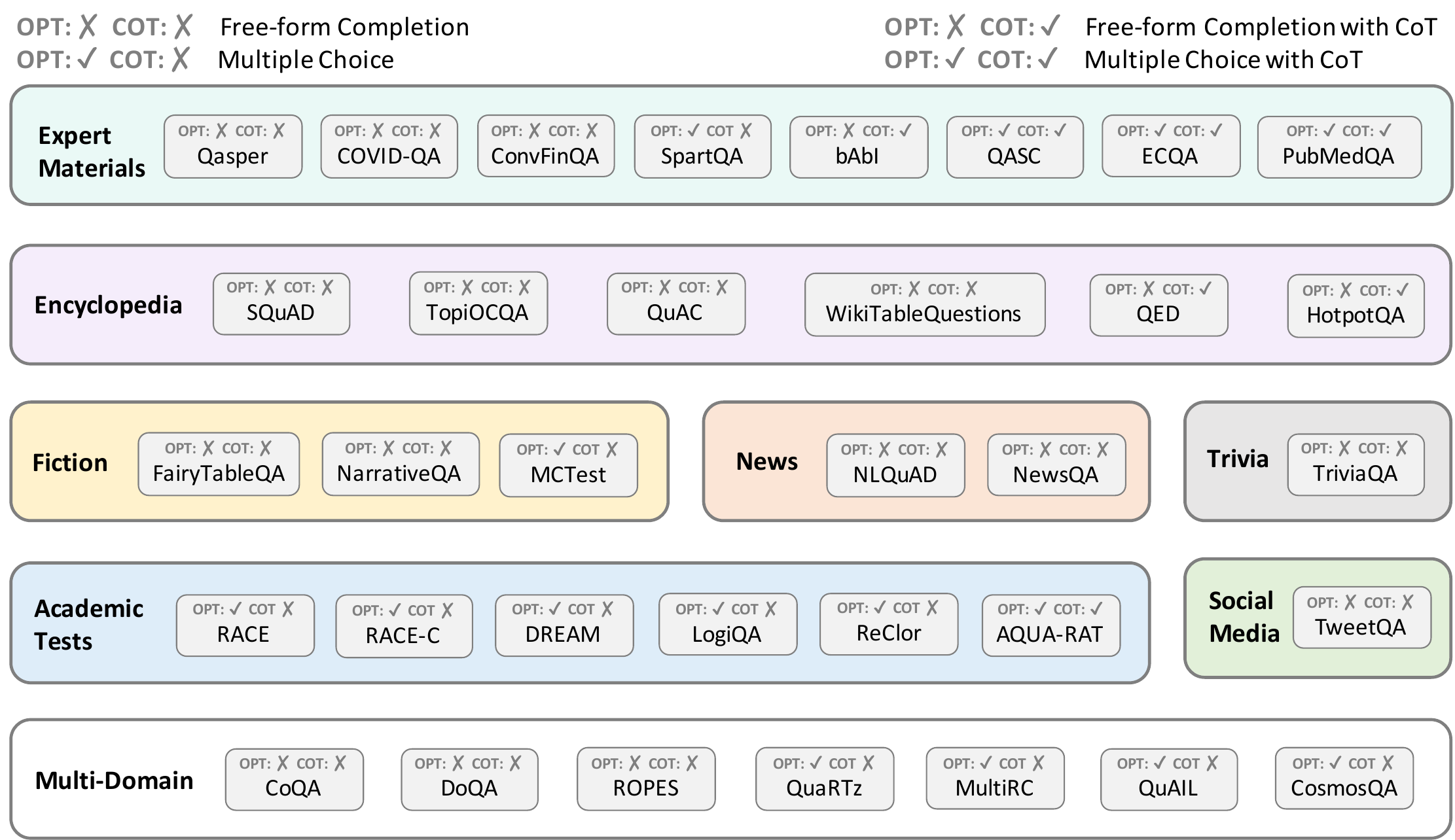}
    \caption{\textbf{Datasets for fine-tuning the instruction synthesizer}, including \citet{qasper, COVIDQA, ConvFinQA, spartqa, bAbI, qasc, ecqa, PubMedQA} in the expert materials domain, \citet{SQuAD, topiocqa, quac, wikitablequestions, qed, hotpotqa} in the encyclopedia domain, \citet{FairytaleQA, narrativeqa, mctest} in the fiction domain, \citet{nlquad, newsqa} in the news domain, \citet{triviaqa} in the trivia domain, \citet{RACE, racec, dream, logiqa, reclor, aquarat} in the academic tests domain, \citet{tweetqa} in the social media domain, and \citet{coqa, doqa, ropes, quartz, multirc, quail, cosmosqa} in the multi-domains sources domain.}

\label{fig:instruction_synthesize_datasets}
\end{figure*}

Figure~\ref{fig:instruction_synthesize_datasets} displays our dataset collection for fine-tuning the instruction synthesizer. For each context in the datasets, we gather all the downstream tasks corresponding to the context, and regard the context as the raw text and the downstream tasks as the instruction-response pairs. For each dataset, we sample a maximum of 10K examples with the highest number of instruction-response pairs, to enhance task diversity while avoiding dataset predominance. Instruction-response pairs covers all the formats defined in~\citep{flancollection}, including free-form completion, multiple-choice, free-form completion with chain-of-thought (CoT;~\citealp{cot}) and multiple-choice with CoT.

\section{Tuning and Inference Settings for Instruction Synthesizer}\label{sec:Tuning and Inference Settings for Instruction Synthesizer}

\paragraph{Data Format} We fill each data example into a specifically designed template to explicitly separate different parts. This facilitates the direct extraction of instruction-response pairs after inference. We use the template {\small \texttt{$<$CON$>$ \{text\} $<$/CON$>$}} to wrap the raw text. As shown in Table~\ref{tab:instruction_prompt_template}, we design different templates for different formats of instructions, and \textbackslash\texttt{n}\textbackslash\texttt{n} is used to connect instruction-response pairs and link them with the raw text. Additionally, we use {\small \texttt{$<$s$>$}} before the beginning of each example and {\small \texttt{$<$/s$>$}} after the end of each example. An $N$-shot example is made by directly concatenating $N$ examples in a sequence. A case of a formatted 2-shot data example for fine-tuning is displayed in Table~\ref{tab:finetune_case}.

\begin{table}[!htb]
\centering
\resizebox{\linewidth}{!}{%
\begin{tabular}{l}
\toprule
\textbf{Instruction Synthesizer Template}                                                                                                                                     \\ \midrule
\multicolumn{1}{l}{\hspace{-0.22cm} {\ul Free-form Completion}} \\
 \begin{tabular}[c]{@{}l@{}} {\small \texttt{$<$QUE$>$ \{instruction\}}} {\small\texttt{$<$ANS$>$ \{response\} $<$/END$>$}}\\\end{tabular}                       \\\midrule[0.01pt]
 \multicolumn{1}{l}{\hspace{-0.22cm} {\ul Multiple Choice}} \\
\begin{tabular}[c]{@{}l@{}} {\small \texttt{$<$QUE$>$ \{instruction\}}}\\ {\small \texttt{Options:} }\\ {\small \texttt{- \{option$_{1}$\}}} \\ {\small \texttt{- \{option$_{2}$\} $<$ANS$>$ \{response\} $<$/END$>$ }} \end{tabular}                                                                    \\\midrule[0.01pt]
\multicolumn{1}{l}{\hspace{-0.22cm} {\ul Free-form Completion with CoT}} \\
\begin{tabular}[c]{@{}l@{}} {\small \texttt{$<$QUE$>$ \{instruction\}}}\\ {\small \texttt{Let's think step by step.} \texttt{$<$ANS$>$ \{CoT\} }} \\{\small \texttt{Therefore, the answer is \{response\} $<$/END$>$ } } \end{tabular}                                                                             \\\midrule[0.01pt]
\multicolumn{1}{l}{\hspace{-0.22cm} {\ul Multiple Choice with CoT}} \\
\begin{tabular}[c]{@{}l@{}} {\small \texttt{$<$QUE$>$ \{instruction\}}}\\ {\small \texttt{Options:} }\\ {\small \texttt{- \{option$_{1}$\}}} \\ {\small \texttt{- \{option$_{2}$\} }} \\ {\small \texttt{Let's think step by step.} \texttt{$<$ANS$>$ \{CoT\}}} \\{\small \texttt{Therefore, the answer is \{response\} $<$/END$>$}} \end{tabular}              \\ \bottomrule
\end{tabular}%
}
\caption{\textbf{Templates for different formats of instruction-response pairs} for tuning and inference of the instruction synthesizer.}
\label{tab:instruction_prompt_template}
\end{table}

\paragraph{Tuning} To constitute a few-shot example for fine-tuning, we concatenate as many formatted examples as possible from the same dataset to match the maximum sequence length. The tuning hyperparameters are in Table~\ref{tab:Hyper-parameters of Fine-tuning the Instruction Synthesizer}. 

\paragraph{Inference} During each round of inference, we concatenate the formatted examples from previous rounds with the formatted raw text of the current round as the input for the instruction synthesizer. Subsequently, the instruction synthesizer generates a sequence of instruction-response pairs. The maximum sequence length for inference corresponds to that of the target LM intended for pre-training. We use the {vLLM}~\citep{vllm} framework for acceleration. It takes about 1 day to synthesize instruction-response pairs for 1B tokens of raw corpora on a single A100-80GB GPU.

\begin{table}[!tb]
\centering
\resizebox{\linewidth}{!}{%
\begin{tabular}{ll}
\toprule 
\textbf{Hyper-Parameter} & \textbf{Assignment}        \\ \midrule
Base model & Mistral-7B-v0.1         \\
Computing infrastructure & 4 A100-80GB GPUs         \\
Run-time                 & 2 days  \\
Epochs         & 5                          \\ 
Batch size              & 16384 tokens                     \\ 
Max sequence length & 4096                      \\ 
Max learning rate   & 5e-6                      \\ 
Optimizer               & Adam                       \\ 
Adam beta weights       & 0.9, 0.95                  \\ 
Learning rate scheduler & cosine              \\ 
Weight decay            & 0.1                       \\ 
Warm-up steps            & 1000                        \\ 
Gradient clipping     & 1.0                    \\ 
Dropout ratio          & 0.1                   \\ \bottomrule
\end{tabular}%
}
\caption{\textbf{Hyper-parameters of fine-tuning the instruction synthesizer.}}
\label{tab:Hyper-parameters of Fine-tuning the Instruction Synthesizer}
\end{table}

\section{LM Evaluation}\label{sec:Language Model Evaluation Setting}
\paragraph{General Models}
We evaluate 0-shot performance on tasks originally formatted as language modeling, including WinoGrande~\citep{winogrande}, PIQA~\citep{piqa} and HellaSwag~\citep{hellaswag}, and 5-shot performance on tasks that are rather challenging and formatted as question-answering, including ARC~\citep{arc}, BoolQ~\citep{boolq}, SIQA~\citep{siqa}, OBQA~\citep{obqa}, and MMLU~\citep{mmlu}. Using the lm-evaluation-harness framework, we report the acc-norm score to follow~\citet{gpt3}.

\paragraph{Domain-Specific Models} We follow the prompting settings of AdaptLLM~\citep{adaptllm}: for biomedicine domain, we evaluate zero-shot performance on PubMedQA~\citep{PubMedQA} and USMLE~\citep{USMLE}, few-shot performance on ChemProt~\citep{ChemProt}, MQP~\citep{MQP} and RCT~\citep{RCT}; for finance domain, we evaluate zero-shot performance on ConvFinQA~\citep{ConvFinQA} and few-shot performance on FPB~\citep{FPB}, FiQA SA~\citep{FiQASA}, Headline~\citep{Headline}, and NER~\citep{ner}.

\begin{table*}[!htb]
\centering
\small
\resizebox{\linewidth}{!}{%
\begin{tabular}{lccccccccc}
\toprule
                   & ARC-e/c                       & BoolQ                   & SIQA            & WG      & PIQA            & OBQA            & HS       & MMLU            \\ \midrule
Total Eval Examples         & 2376/1172            & 3270                     & 1954            & 1267            & 1838            & 500            & 10042            & 14042            \\
Contam in Raw Corpora & 5/3        & 144            & 0            & 0   & 3           & 0            & 4   &20\\
Contam in Ins-Aug Corpora & 5/4        & 144            & 0            & 0   & 3           & 0            & 4   &22 \\

Contam in Synthesized Pairs & 0/1            & 0                   & 0            & 0            & 0           & 0    & 0   & 2            \\\bottomrule
\end{tabular}%
}
\caption{\textbf{Data contamination analysis} of raw corpora, instruction-augmented corpora and the synthesized instruction-response pairs. ``WG'' and ``HS'' represent WinoGrande and HellaSwag, respectively.}
\label{tab:Data Contamination Analysis}
\end{table*}

\begin{table}[!htb]
\centering
\small
\resizebox{\columnwidth}{!}{
\begin{tabular}{lccc}
\toprule
 & Accuracy  & Relevance & \# Category \\ \midrule
General   & 75.5 & 87.5  & 51  \\
BioMed.   & 81.0 & 97.0  & 21  \\
Finance   & 73.5 & 88.0  & 39  \\ \bottomrule
\end{tabular}
}
\caption{\textbf{Human evaluation of response accuracy, context relevance, and number of task categories} on the instruction-augmented corpora.}
\label{tab:human analysis}
\end{table}

\section{Data Contamination Analysis}\label{sec: Data Contamination Analysis}

We measure cross-contamination between the evaluation datasets and the training data using the sub-string match method described in~\citet{gpt4}: an evaluated example is considered contaminated if a sub-string of it appears in the training data. Table~\ref{tab:Data Contamination Analysis} shows:
\begin{itemize}[leftmargin=*]
\itemsep0em 
\item \textit{Total Eval Examples}: The number of all evaluated examples in each dataset.
\item \textit{Contam in Raw Corpora}: The number of contaminated examples in the raw corpora.
\item \textit{Contam in Ins-Aug Corpora}: The number of contaminated examples in the instruction-augmented corpora, which includes the raw corpora and the synthesized instruction-response pairs.
\item \textit{Contam in Synthesized Pairs}: The number of contaminated examples introduced by synthesized pairs, calculated by subtracting the number of contaminated examples in the raw corpora from those in the instruction-augmented corpora.
\end{itemize}

The results indicate the synthesized pairs introduce minimal contamination to the training data.

\begin{table}[!htb]
\centering
\small
\resizebox{\columnwidth}{!}{
\begin{tabular}{ccc}
\toprule
Coverage   & Coverage (multi-domain) & Overlap \\ \midrule
86.8   & 77.8    &  84.9  \\\bottomrule
\end{tabular}
}
\caption{\textbf{Domain coverage and overlap} between the raw text and the synthesized instruction-response pairs.}
\vspace{-5pt}
\label{tab:Domain Coverage and Overlap}
\end{table}

\begin{table*}[!htb]
\centering
\small
\resizebox{\linewidth}{!}{%
\begin{tabular}{lccccccccc}
\toprule
          & Encyclo & Fiction & Academic & Trivia & News    & Expert & Social & Code   \\\midrule
Fine-tune Data & 22.2      & 11.1 & 22.2        & 3.7 & 7.4  & 29.6          & 3.7       & 0.0 \\
Raw Corprora                  & 5.8       & 9.6  & 3.5         & 0.0 & 20.3 & 42.8          & 14.7      & 3.3 \\
Synthesized Pairs             & 5.8       & 11.8 & 3.3         & 0.1 & 18.8 & 46.0          & 11.1      & 3.1\\\bottomrule

\end{tabular}%
}
\caption{\textbf{Domain distribution} of fine-tuning data for the instruction synthesizer, raw corpora and synthesized instruction-response pairs. ``Encyclo", ``Academic", ``Expert" and ``Social" represent Encyclopedia, Academic Tests, Expert Materials and social media domains, respectively.}
\label{tab: Domain Distribution}
\end{table*}

\section{Human Evaluation on Instruction-Augmented Corpora}\label{sec: Human Evaluation on Instruction-Augmented Corpora}
We conduct human evaluation to analyze the instruction-augmented corpora from the following aspects:
\begin{itemize}[leftmargin=*]
\itemsep0em 
\item \textit{Response Accuracy}: A binary score indicating whether the response is accurate based on the instruction and context, where 1 means accurate and 0 means inaccurate. We report the average score of all responses.
\item \textit{Context Relevance}: A binary score indicating whether the instruction-response pair is relevant to the context of the raw text, where 1 means relevant and 0 means irrelevant. We report the average score of all instruction-response pairs.
\item \textit{\# Task Category}: The evaluator categorizes each instruction-response pair using a predefined list of task categories from \citet{supernaturalinstructions}. We report the number of different categories of all the instruction-response pairs to show diversity.
\end{itemize}

From the results in Table~\ref{tab:human analysis}, the synthesized instruction-response pairs span 51 different task categories, with over 85\% relevance to the context and 70\% response accuracy.

\section{Domain Distribution Analysis}\label{sec:Analysis on Domain Diversity}
We analyze domain distribution to evaluate the effectiveness of our instruction synthesizer in generating instruction-response pairs closely aligned with the domain of the given raw context.
\paragraph{Domain Coverage and Overlap} For each instruction-augmented text, we calculate the following scores and report the average on all the instruction-augmented texts:

\begin{itemize}[leftmargin=*]
\itemsep0em 
\item \textit{Domain Coverage}: The ratio of text domains included in the instruction domains to all text domains.
\item \textit{Domain Coverage (multi-domain only)}: We specifically compute domain coverage for the cases where a raw text contains multiple domains.
\item \textit{Domain Overlap}: The overlap of raw text domains and instruction domains divided by the union of raw text and instruction domains.
\end{itemize}

As shown in Table~\ref{tab:Domain Coverage and Overlap}, the synthesized instruction-response pairs cover most of the domains in the raw text, with a high domain overlap with the raw text. For the texts containing more than one domain, our instruction synthesizer generates, on average, 5 instruction-response pairs per raw text, with each pair potentially covering a different domain. According to the \textit{domain coverage (multi-domain only)}, when a single raw text includes multiple domains, our instruction synthesizer can effectively generate instruction-response pairs that cover most of the text domains.

\paragraph{Domain Distribution} We analyze domain distributions of the following sources: 
\begin{itemize}[leftmargin=*]
\itemsep0em 
\item Fine-tuning data for the instruction synthesizer.
\item Raw pre-training corpora~\citep{refinedweb}.
\item Synthesized instruction-response pairs based on the raw pre-training corpora.
\end{itemize}

As shown in Table~\ref{tab: Domain Distribution}, despite the domain distributions of fine-tuning data and raw corpora being very different, the synthesized pairs closely follow the domain distribution of the raw corpora.

\begin{table*}[!b]
\centering
\resizebox{\linewidth}{!}{%
\begin{tabular}[c]{@{}m{\linewidth}@{}}
\toprule
{\small \texttt{$<$s$>$}} {\small \texttt{$<$CON$>$}} Our school life is very interesting! My friends and I study hard at school. And we are good at our lessons. We are very happy. We have lots of time for our hobbies. My classmates all want to go to different clubs. Helen wants to join the Reading Club. She loves reading books. The Reading Club meets every Wednesday at three thirty. Lily enjoys dancing. She wants to join the Dancing Club. It meets on Mondays at four thirty. There's also an Art Club. It meets on Fridays at four o'clock. Nick doesn't want to join the Art Club. He doesn't like drawing. He thinks it is too difficult for him . Nick likes playing computer games. He wants to join the Computer Club. It meets every Thursday at three forty-five. Mike loves sports. He wants to join the football team. They play football every Monday at three thirty. I want to join the Music Club. I like listening to music with my friends. The Music Club meets on Tuesday at three fifteen. {\small \texttt{$<$/CON$>$}}
\\
\\
{\small \texttt{$<$QUE$>$}} What club does Helen like? {\small \texttt{$<$ANS$>$}} Helen likes the reading club. {\small \texttt{$<$/END$>$}}
\\
\\
{\small \texttt{$<$QUE$>$}} How many friends does the story teller describe? {\small \texttt{$<$ANS$>$}} I have four friends. {\small \texttt{$<$/END$>$}}
\\
\\
{\small \texttt{$<$QUE$>$}} Are you and your friends smart? {\small \texttt{$<$ANS$>$}} unknown {\small \texttt{$<$/END$>$}} {\small \texttt{$<$/s$>$}}{\small \texttt{$<$s$>$}} {\small \texttt{$<$CON$>$}} Billy and Sara are brother and sister. They went to the beach with their family last July for a week, and had the best time ever! On Monday, Billy and Sara wanted to build a giant sandcastle. They invited their new friends Jack and Jane to help build the sandcastle. Jack and Jane had a house on the beach, so they were really good when it came to building sandcastles. They hoped that they could make the sandcastle taller than themselves, but they soon found they needed more help. They asked their cousin Joey to help them build the biggest sandcastle in the world! Joey wasn't the friendliest cousin in the world, but to Billy and Sara's surprise, Joey was happy to help build the sandcastle. Billy, Sara, Jake, Jane and Joey had spent the whole day building the sandcastle, and finally, right before dinner time, they completed it. The sandcastle was huge! It had a river around the castle, and even a bridge to cross the river. It even had a flag at the top, and a wall that went around the castle too! They were so happy!\\
\\
The rest of the week at the beach was a lot of fun for Billy and Sara. On Tuesday, they went for ice cream. Sara's ice cream fell and dripped all the way down to her tummy, but Billy gave her some of his. On Wednesday, they watched the fireworks at night. On Thursday, they went swimming all day long, moving like worms in the water. On Friday, they had to go back home. They were sad, so they started counting down the days until next year at the beach! {\small \texttt{$<$/CON$>$}}
\\
\\
{\small \texttt{$<$QUE$>$}} how do billy and Sara know each other? {\small \texttt{$<$ANS$>$}} Billy and Sara are brother and sister. {\small \texttt{$<$/END$>$}}\\
\\
{\small \texttt{$<$QUE$>$}} Did they do something yesterday? {\small \texttt{$<$ANS$>$}} no. {\small \texttt{$<$/END$>$}}\\
\\
{\small \texttt{$<$QUE$>$}} When did they do something? {\small \texttt{$<$ANS$>$}} last July {\small \texttt{$<$/END$>$}}\\
\\
{\small \texttt{$<$QUE$>$}} What did they do?
{\small \texttt{$<$ANS$>$}} They went to the beach {\small \texttt{$<$/END$>$}} {\small \texttt{$<$/s$>$}}\\ \bottomrule
\end{tabular} %
}
\caption{\textbf{An example of a sequence for fine-tuning the instruction synthesizer.} This sequence contains two examples, both from the CoQA dataset \citep{coqa}, constituting a 2-shot example.}
\label{tab:finetune_case}
\end{table*}

\begin{table*}[!htb]
\centering
\begin{tabular}{llll}
\toprule
\textbf{Hyper-Parameter} & \multicolumn{2}{l}{\textbf{Pre-Train from Scratch}} & \textbf{Continual Pre-Train} \\ \midrule
Parameters               & 500M                      & 1.3B                   & 8B                       \\
Hidden size              & 1536                      & 2048                   & 4096                     \\
Intermediate size        & 4320                      & 8192                   & 14336                    \\
Max Position Embeddings  & 2048                      & 2048                   & 8192                     \\
Num attention heads      & 24                        & 32                     & 32                       \\
Num hidden layers        & 16                        & 20                     & 32                       \\
Num key value heads      & 24                        & 8                      & 8                        \\
Rope theta               & 10000                     & 10000                  & 500000                   \\
Vocab Size               & 32000                     & 32000                  & 128256                   \\
Tokenizer                & Mistral                   & Mistral                & Llama3                   \\
Computing infrastructure & 8 A100-80GB GPUs          & 8 A100-80GB GPUs       & 4 A100-80GB GPUs         \\
Run-time                 & 5 days                    & 10 days                & 1 day                    \\
Train steps              & 200K                      & 100K                   & 4K                       \\
Batch size               & 0.5M tokens               & 1M tokens              & 0.25M tokens             \\
Max Sequence Length      & 2048                      & 2048                   & 4096                     \\
Max Learning Rate        & 3e-4                  & 2e-4               & 1e-5                 \\
Optimizer                & Adam                      & Adam                   & Adam                     \\
Adam beta weights        & 0.9, 0.95                 & 0.9, 0.95              & 0.9, 0.95                \\
Learning rate scheduler  & cosine                    & cosine                 & cosine                   \\
Weight decay             & 0.1                       & 0.1                    & 0.1                      \\
Warm-up steps            & 2000                      & 2000                   & 1000                     \\
Gradient clipping        & 1                         & 1                      & 1                        \\
Dropout ratio            & 0.1                       & 0.1                    & 0.1                      \\ \bottomrule
\end{tabular}%
\caption{\textbf{Hyper-parameters of pre-training from scratch and continual pre-training.}}
\label{tab:Hyper-parameters of General Pre-training From Scratch}
\end{table*}

\begin{table*}[!htb]
\centering
\resizebox{\linewidth}{!}{%
\begin{tabular}{lllcccccccccc}
\toprule
            & \# Param. & \# Token & ARC-e/c                   & BoolQ      & SIQA   & WG     & PIQA   & OBQA   & HS     & MMLU   \\\midrule
Instruct PT & 500M      & 100B      & 54.8/27.4               & 62.0    & 47.2   & 54.8   & 69.9   & 30.8   & 47.3   & 25.3   \\
GPT-2 & 774M      & -         & 53.8/24.9              & 62.1     & 45.5   & 54.5   & 69.3   & 30.6   & 45.3   & 25.5   \\
Pythia      & 1B        & 300B      & 59.0/28.8            & 61.6      & 46.3   & 52.6   & 69.3   & 32.6   & 47.2   & 26.1   \\
BLOOM       & 1.1B      & 341B         & 52.3/28.3            & 61.5      & 45.9   & 52.7   & 67.2   & 28.6   & 43.0   & 26.6   \\\midrule
Instruct PT & 1.3B      & 100B      & 60.5/30.9               & 62.2     & 49.2   & 55.9   & 73.6   & 33.4   & 54.3   & 27.3   \\
OPT         & 1.3B      & 300B      & 60.1/31.1   & 62.4     & 48.4   & 58.2   & 71.0   & 34.0   & 53.8   & 25.1   \\
GPT-2    & 1.5B      & -         & 60.2/29.6            & 63.5     & 47.3   & 56.2   & 70.5   & 33.2   & 50.8   & 26.3   \\
BLOOM      & 3B      & 341B      & 63.1/35.3   & 62.2     & 48.8   & 57.4   & 70.5   & 33.0   & 54.6   & 25.9  \\ \bottomrule
\end{tabular}%
}
\caption{\textbf{Comparison between our pre-trained models and other open-source models}~\citep{gpt2, pythia, Bloom, opt} on general benchmarks. ``WG'' and ``HS'' represent WinoGrande and HellaSwag, respectively.}
\label{tab:general_comprison}
\end{table*}

\begin{table*}[!htb]
\centering
\resizebox{\linewidth}{!}{%
\begin{tabular}[c]{@{}m{\linewidth}@{}}
\toprule
Not a writer, a writer wannabe, editor, lit maj, or pretend literary critic. Just an avid reader/listener. My ratings are opinion only.\\
I love all genres of books. However, when I listen to audio books as I clean, garden, drive they are better with a lot of heat!\\
"Laborious"\\
This might have been a bit more tolerable if narrator was better. I am happy to say that I did finish the book but it just seemed to go and on. Like other listeners the book itself reminded me of a bad TV show. Not horrible but of all the books I have listened to this is just bearly average.\\
\\
Problem: Pick your answer from:\\
a). They didn't like the genre.;\\       
b). They did n't have enough time to read it.;\\
c). They did n't like the author.;\\            
d). They did n't like the narrator.;\\       
Q: What may be the reason for them not finishing the book?\\
\\
Answer: d).\\
\\
Customer Web Interaction: Fundamentals and Decision Tree From Virtual Communities\\
Authors\\
Enrico Senger, Sandra Gronover, and Gerold Riempp, University of St. Gallen\\
Abstract\\
In order to utilise the new possibilities of Internet technology efficiently, many companies invest considerable sums in the development of communication channels to customers. In this context, the often-quoted objective of cost saving per interaction appears to be questionable, since new communication media have not been able to fully substitute the existing systems. Costs are therefore more likely to rise than drop. The following article discusses potentials, criteria, conditions and consequences related to the use of computer-mediated environments for customer interaction. The objective is to derive recommendations for action in respect of a context-dependent
support, especially by means of web collaboration and self-service-options.\\
Download Customer Web Interaction: Fundamentals and Decision Tree\\
\\
Problem: Pick your answer from:\\
a). It can be edited.;\\
b). It can be read offline.;\\
c). It can be read online.;\\
d). It can be used offline.;\\
Q: What may happen after the download?\\
\\
Answer: c).\\ \bottomrule
\end{tabular} %
}
\caption{\textbf{A case of a 2-shot example in the general instruction-augmented corpora.}}
\label{tab:pt_case_general}
\end{table*}

\begin{table*}[!htb]
\centering
\resizebox{\linewidth}{!}{%
\begin{tabular}[c]{@{}m{\linewidth}@{}}
\toprule
Read this article and answer questions\\
\# Correlation between increased airway responsiveness and severity of pulmonary edema.\\
To determine whether the severity of the pulmonary edema in sheep models of cardiogenic and non-cardiogenic pulmonary edema correlate with concomitant alterations in airway responsiveness using three separate measures of pulmonary edema: post-mortem wet-to-dry lung weight ratio (W/D), chest radiograph (CXR) scores, and small airway wall area. Cardiogenic pulmonary edema was induced by increasing left atrial pressure (increase PLA) and non-cardiogenic pulmonary edema was induced by intravenous administration of Perilla ketone (PK). \textbf{(...)}\\
\\
Does increased airway responsiveness correlate with pulmonary edema severity in sheep?\\
--\\
Let's think first: Increased airway responsiveness correlates with severity of pulmonary edema in sheep... So the answer is [Yes]\\
\\
Read this article and answer questions\\
\# Immobilization and bioactivity evaluation of FGF-1 and FGF-2 on powdered silicon-doped hydroxyapatite and their scaffolds for bone tissue engineering.\\
Fibroblast growth factors (FGFs) are polypeptides that control the proliferation and differentiation of various cell types including osteoblasts. FGFs are also strong inducers of angiogenesis, necessary to obtain oxygen and nutrients during tissue repair. \textbf{(...)}\\
\\
Do immobilization of fibroblast growth factors 1 and 2 on silicon-doped hydroxyapatite scaffolds for bone repair?\\
--\\
Let's think first: This study demonstrated the efficient immobilization of FGF-1 and FGF-2 on Si-HA and Si-HA scaffolds, retaining their biological activity on osteoblasts. Thus, these FGF/scaffolds may be useful in bone tissue engineering applications... So the answer is [Yes]\\
\\
Read this article and answer questions\\
\# Phytotoxicity, cytotoxicity and genotoxicity evaluation of organic and inorganic pollutants rich tannery wastewater from a Common Effluent Treatment Plant (CETP) in Unnao district, India using Vigna radiata and Allium cepa.\\
The leather industry is a major source of environmental pollution in India. The wastewater generated by leather industries contains very high pollution parameters due to the presence of a complex mixture of organic and inorganic pollutants even after the treatment at a Common Effluent Treatment Plant (CETP) and disturbs the ecological flora and fauna. The nature, characteristics and toxicity of CETP treated wastewater is yet to be fully elucidated. Thus, this study aims to characterize and evaluate the toxicity of CETP treated tannery wastewater collected from the Unnao district of Uttar Pradesh, India. In addition to measuring the physico-chemical parameters, the residual organic pollutants was identified by GC-MS analysis and phytotoxicity, cytotoxicity and genotoxicity of the treated wastewater was evaluated using Vigna radiata L. and Allium cepa L. \textbf{(...)}\\
\\
Is common effluent treatment plant wastewater safe for the environment?\\
--\\
Let's think first: The present study revealed the presence of high levels of various pollutants in CETP treated tannery wastewater. Moreover, the toxicity assessment showed the phytotoxic and genotoxic nature of the wastewater which suggests that this wastewater cannot be directly discharged into the environment without any further treatment... So the answer is [No]\\ \bottomrule
\end{tabular} %
}
\caption{\textbf{A case of a 3-shot example in the instruction-augmented corpora for biomedicine domain.} Certain portions are omitted for brevity and are represented as \textbf{(...)}}
\label{tab:pt_case_med}
\end{table*}

\begin{table*}[!htb]
\centering
\resizebox{\linewidth}{!}{%
\begin{tabular}[c]{@{}m{\linewidth}@{}}
\toprule
Answer questions based on this article:\\     
Once the MOASS is truly over would anyone like an AMA with DFV AND RC? I would love to learn what went on through their minds and the events all the way from 2019 to post-MOASS.\\
They must be dying to talk about all the things that went on (but couldn't because of all the potential controversy and lawsuits that can be had) and apes would love to get the official explanation on the cryptic, and some not so cryptic tweets from DFV and RC. Edit: it may be obvious but it's just an opinion of mine on to see what they may have to say. If it does somehow gain enough traction, we would respectfully ask them if they're interested. If not, no AMA. Simple as that. I've been thinking what we should do is once the squeeze is over let it die down a bit and then we should start a gmecon or something similar. I wanted to right a post about it but my karma is too low so if someone else wants to put it out there and see what people think that would be great. Personally I'm in this stock for life and would love an annual event where we could all meet up and have in person Q and A's with RC, DFV and others, even someone like Jordan Belfort to hype up the apes after we take our tendies. also would be good to see all gamestops ideas for the future. Just a thought hope there's some way we could make this happen.\\
\\
question below:\\
What might happen after the MOASS?\\
answer below:\\     
People will want an AMA with DFV and RC\\
\\
question below:\\
What might happen if they did an AMA with DFV and RC?\\
answer below:\\
They would ask questions about the cryptic tweets\\
\\
Answer questions based on this article:\\
Pixar's `Lightyear' snares \$51 million in domestic opening\\
Pixar's ``Lightyear'' rocketed to a \$51 million domestic opening, the best performance of an animated feature since the pandemic began. Internationally, the Disney film tallied \$34.6 million in ticket sales, bringing its global haul to \$85.6 million. The animated film's performance, while strong for a pandemic release, fell short of expectations. Box office analysts had foreseen ``Lightyear'' bringing in between \$70 million and \$85 million domestically. Expectations were high because the last two films in the Toy Story franchise both opened to more than \$100 million in ticket sales, according to data from Comscore. ``Toy Story 4'' in 2019 topped \$120 million in its domestic debut and ``Toy Story 3'' generated more than \$110 million during its opening 2010. ```Lightyear' had a great deal of potential on paper, but a number of factors resulted in this very rare box office misfire for a Pixar release,'' said Shawn Robbins, chief media analyst at BoxOffice.com. It's unclear if tough box office competition with Universal's ``Jurassic World: Dominion,'' which generated \$58.6 million over the weekend, and Paramount and Skydance's ``Top Gun: Maverick,'' which secured another \$44 million, was the reason for ``Lightyear's'' smaller-than-expected opening or if consumers were confused about the film release. After all, there has not been a theatrical release of a Pixar film since 2020's ``Onward.'' \textbf{(...)}\\
\\
question below:\\
What is the main point of the article?\\
answer below:\\
Lightyear fell short of expectations\\
\\
question below:
What is the author's opinion of why the movie had a smaller than expected opening?\\
answer below:\\
It had tough box office competition\\ \bottomrule
\end{tabular} %
}
\caption{\textbf{A case of a 2-shot example in the instruction-augmented corpora for finance domain.} Certain portions are omitted for brevity and are represented as \textbf{(...)}}
\label{tab:pt_case_finance}
\end{table*}

\end{document}